\documentclass[letterpaper,10pt,conference]{ieeeconf}

\IEEEoverridecommandlockouts
\usepackage{amsmath}
\usepackage{amssymb}
\usepackage{booktabs}
\usepackage{cite}
\usepackage{cuted}
\preCutedStrip={\vskip -22pt}
\usepackage{capt-of}
\usepackage{graphicx}
\usepackage{multirow}
\usepackage{stfloats}
\usepackage{xcolor}
\usepackage{url}
\makeatletter
\let\NAT@parse\undefined
\makeatother
\usepackage[colorlinks=true,allcolors=black]{hyperref}

\graphicspath{{figures/}}

\newif\ifdraft
\draftfalse

\title{\LARGE \bf
AdaTempo: Learning Shared Relative Tempo from Demonstrations\\
for Faster Robot Manipulation
}

\author{\authorblockN{Jiale Cao$^{1}$, Yike Niu$^{2}$, Zhengrong Xue$^{2}$, and Huazhe Xu$^{2,*}$}%
\authorblockA{$^{1}$Fudan University \qquad $^{2}$Tsinghua University}%
\thanks{$^{*}$Corresponding author}%
\thanks{Contact: caojl23@m.fudan.edu.cn, huazhe\_xu@mail.tsinghua.edu.cn.}}

\begin{document}

\maketitle
\thispagestyle{empty}
\pagestyle{empty}

\begin{strip}
    \centering
    \makebox[\textwidth][c]{%
        \includegraphics[width=1.08\textwidth]{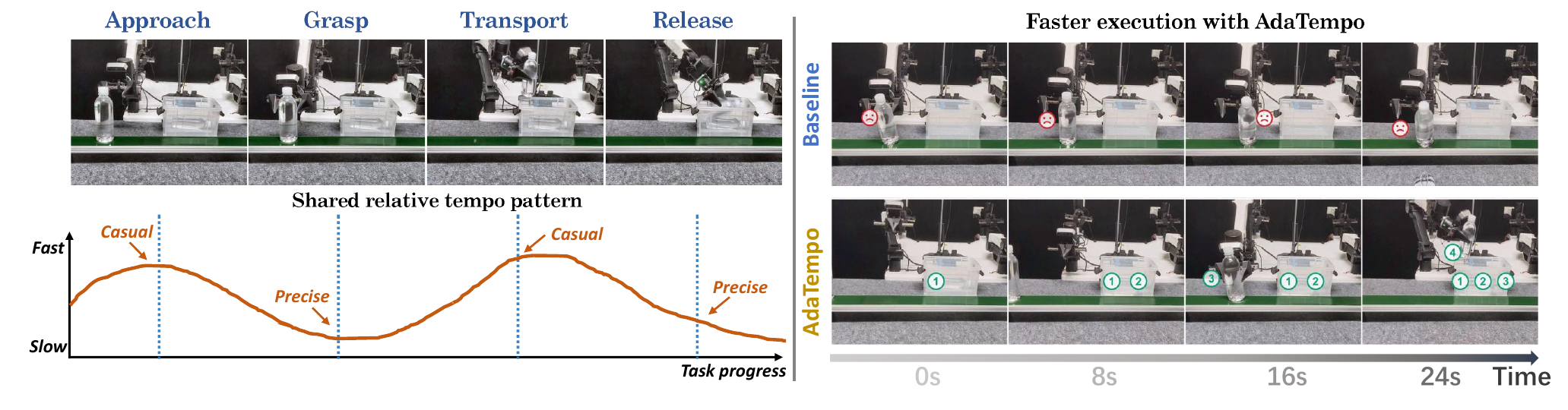}%
    }
    \captionof{figure}{\textbf{Overview.} Left: Demonstrations exhibit shared phase-dependent relative-tempo patterns, with faster motion in casual phases and slower motion in precision-critical phases. \textbf{AdaTempo} exploits this structure to enable non-uniform acceleration across task progress. Right: On a fast conveyor-belt task, AdaTempo keeps pace with the production line and successfully grasps fast-moving objects that the ACT baseline frequently misses.}
    \label{fig:teaser}
\end{strip}

\begin{abstract}
Visuomotor policies trained via imitation learning
often inherit the unnecessarily slow timing of teleoperated
demonstrations. Yet uniform speedup is unreliable because
different phases of a manipulation task tolerate acceleration
differently. In this work, we introduce AdaTempo, a self-supervised method that accelerates visuomotor policies by exploiting shared relative-tempo structure in demonstrations. AdaTempo establishes phase correspondence, aggregates the aligned relative tempo into a consensus, and maps it to a continuous speedup profile used to resample demonstrations into accelerated training trajectories. Training
standard policies such as ACT or Diffusion Policy on these
resampled trajectories directly embeds the desired tempo in the
learned behavior, without runtime tempo selection or online
retiming. Extensive evaluations show that AdaTempo achieves
up to a $3.57\times$ speedup and yields a stronger success--speed trade-off than the original policies and representative acceleration baselines.

\end{abstract}

\section{Introduction}
\label{sec:introduction}

Visuomotor imitation learning has become a practical paradigm for acquiring
complex manipulation skills from teleoperated
demonstrations~\cite{zhao2023act,chi2023diffusion,ze2024dp3}. Recent
large-scale policies have further improved generalization across tasks and
robot embodiments~\cite{ghosh2024octo,kim2025openvla,black2024pi0}. Yet standard imitation
objectives optimize what actions to reproduce, not how quickly they should be
executed. Since teleoperated demonstrations are often slow and cautious, the
resulting policies tend to inherit this conservative execution
timing~\cite{guo2025demospeedup,arachchige2025sail,
yuan2025speedtuning,kim2026race}.

Removing this inefficiency is not as simple as downsampling the demonstrated
trajectories by a constant rate. Different phases of a manipulation task
tolerate acceleration differently. Uniform acceleration can skip important
intermediate states, alter the closed-loop state distribution, or request
motions that the controller cannot track over the shortened time
horizon~\cite{arachchige2025sail,kim2026race,nam2025speedaug}. Effective
speedup therefore requires determining both how much to accelerate a task and
how to distribute that acceleration over its progress.

Recent work has made substantial progress in accelerating learned manipulation
policies beyond demonstration
speed~\cite{guo2025demospeedup,yuan2025speedtuning,arachchige2025sail,
kim2026race,nam2025speedaug}. However, existing methods typically determine
how acceleration should vary using policy-derived proxies,
interaction-based optimization, or additional runtime mechanisms, requiring auxiliary machinery during training or deployment. This motivates a
central question: can demonstrations themselves provide a direct and
fine-grained supervisory signal for phase-dependent speedup?

As illustrated in Fig.~\ref{fig:teaser}, our key observation is that
demonstrations may differ in absolute speed yet
tend to exhibit similar relative-tempo patterns over task progress, consistent
with prior observations of approximate relative-timing invariance in skilled
movement~\cite{schmidt1985invariance}. Rather than absolute speed, we focus on
relative tempo: within each demonstration, which phases are executed relatively
faster or slower? When corresponding phases exhibit similar relative tempo
across demonstrations, this consistency suggests phase-specific task
characteristics. Phases consistently executed more cautiously, such as
grasping or insertion, may tolerate less temporal compression, while relatively
fast phases such as free-space transport offer greater acceleration potential.

Based on this insight, we introduce \textbf{AdaTempo}, a self-supervised method
for learning cross-demonstration relative tempo to accelerate visuomotor policies. AdaTempo aligns demonstrations with different durations, constructs
episode-relative tempo profiles at corresponding task phases, and aggregates
them into a cross-demonstration consensus while downweighting phases with high
disagreement. It maps this consensus to a continuous speedup profile used
for smooth virtual-time resampling of the original trajectories. A standard visuomotor policy, such as ACT or Diffusion Policy (DP), is then trained on the
accelerated trajectories. The desired tempo is thus embedded directly in the
learned behavior, requiring no runtime tempo-selection module, online retiming,
or changes to the control interface at deployment.

Extensive simulation and real-world evaluations show that AdaTempo achieves
up to a $3.57\times$ speedup over demonstrations and yields a stronger
success--speed trade-off than the original policies and representative
acceleration baselines. Our contributions are:
\begin{itemize}
    \item We identify shared relative tempo across demonstrations as a self-supervised cue for phase-dependent speedup, revealing recurring phase-wise timing structure beyond demonstrator-specific absolute speed.
    \item We develop AdaTempo, which phase-aligns demonstrations, converts their relative-tempo consensus into a continuous speedup profile for offline virtual-time resampling, without auxiliary models or additional mechanisms at deployment.
    \item We validate AdaTempo across 11 simulated and real-world manipulation tasks with ACT or DP, achieving up to a \(3.57\times\) speedup and a stronger success--speed trade-off than representative acceleration baselines.
\end{itemize}

\section{Related Work}
\label{sec:related-work}

\subsection{Visuomotor Imitation Learning}

Modern visuomotor policies learn temporally coherent manipulation behaviors
through action chunking, generative sequence modeling, and increasingly
large-scale multi-task training~\cite{shridhar2022peract,zhao2023act,
chi2023diffusion,ze2024dp3,brohan2022rt1,zitkovich2023rt2,ghosh2024octo,
kim2025openvla,liu2025rdt,black2024pi0,oneill2024openx,
khazatsky2024droid}. These methods typically train on the temporal sampling
provided by demonstrations and can therefore reproduce not only demonstrated
motions but also their execution tempo~\cite{guo2025demospeedup,
arachchige2025sail,yuan2025speedtuning,kim2026race}. Orthogonal work reduces
policy inference latency through distillation, efficient generative modeling,
or asynchronous action generation~\cite{prasad2024consistency,
lu2024manicm,pertsch2025fast,black2025rtc}. AdaTempo instead targets physical task execution time by temporally resampling demonstrations before policy training.

\subsection{Acceleration of Imitation Learning Execution}

Existing approaches accelerate manipulation policies either by modifying policy execution at runtime or by altering demonstrations and training targets before deployment.

\paragraph{Runtime policy-execution acceleration}
Runtime methods advance within action chunks, select speed multipliers, or
retime commands based on motion complexity, tracking error, latency, or
physical reachability~\cite{park2024pte,yuan2025speedtuning,
arachchige2025sail,kim2026race,wu2026speeduppatch}. Related approaches expose
speed explicitly through variable-frequency conditioning or continuous action
parameterizations that can be adjusted at
deployment~\cite{masuya2025vfil,jing2026tempovla,han2026bspline}. While
effective for adapting frozen policies or enabling on-demand speed control,
these approaches require additional mechanisms to select execution
speed at runtime. Some, such as SpeedTuning, further learn speed selection
through reinforcement learning and environment
interaction~\cite{yuan2025speedtuning}.

\paragraph{Demonstration- and training-side acceleration}
Another family of methods accelerates policy execution by modifying the temporal structure of demonstrations or training targets before deployment. DemoSpeedup estimates conditional action entropy with a proxy generative policy
and assigns different downsampling factors to the resulting trajectory
segments~\cite{guo2025demospeedup}. Semantic approaches instead use
vision--language reasoning, gripper--object relations, or propagated stage
labels to identify regions that can be accelerated, while motion-frequency
cues have been used to remove temporally redundant
frames~\cite{kim2025espada,sanchez2026volt,dai2026skip}. AutoSpeed optimizes
over smoothly speed-transformed action targets using a prediction-error and
horizon criterion, whereas SpeedAug combines variable-tempo augmentation with
reinforcement-learning refinement~\cite{hu2026autospeed,nam2025speedaug}. However, these methods typically rely on auxiliary signals or objectives to
determine where and how much to accelerate. In contrast, AdaTempo derives a
continuous speedup signal directly from relative-tempo structure
shared across demonstrations and incorporates it through temporally resampled training data.

\subsection{Temporal Correspondence and Relative Tempo}

Temporal alignment enables comparison between demonstrations that follow
similar task progress with different durations. Dynamic time warping (DTW)
constructs a monotone correspondence between sequences~\cite{sakoe1978dtw}
and has been used to align demonstrations before extracting or transferring
skill structure~\cite{braglia2024arclength}. In acceleration work, ESPADA uses
DTW to propagate semantic stage labels from an annotated trajectory to other
demonstrations~\cite{kim2025espada}. Complementary evidence from motor-control
studies indicates that skilled movements can retain relative timing and
normalized velocity structure under changes in execution
speed~\cite{schmidt1985invariance,summers1975timing,shapiro1981gait,
atkeson1985kinematic}. This observation motivates comparing demonstrations in
a normalized temporal coordinate rather than by their absolute execution
speeds. AdaTempo uses DTW not to propagate predefined labels, but to establish
correspondences across demonstrations, enabling relative tempo to be compared
at matched task progress. Recurring cross-demonstration relative-tempo
structure is then used to derive a continuous speedup profile.

\section{AdaTempo}
\label{sec:method}

\begin{figure*}[t]
    \centering
    \includegraphics[width=\textwidth]{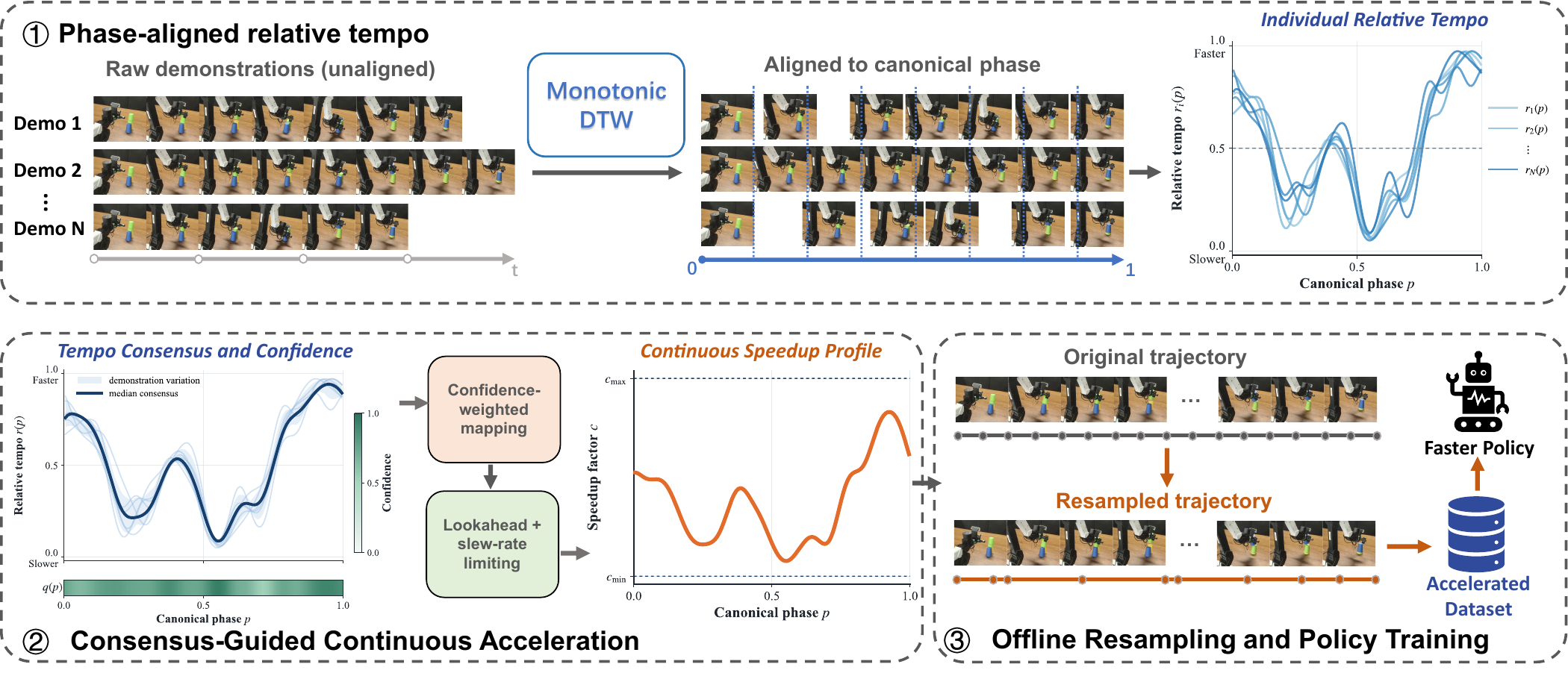}
    \caption{\textbf{Pipeline.} Demonstrations with different durations are
    aligned to a canonical phase. Their relative-tempo consensus and
    agreement determine a continuous speedup profile, which resamples the
    demonstrations for standard visuomotor policy training.}
    \label{fig:pipeline}
\end{figure*}

Let $\mathcal{D}=\{\tau_i\}_{i=1}^{N}$ be a set of $N$ collected demonstrations, where
$\tau_i=\{(o_{i,t},a_{i,t})\}_{t=0}^{T_i-1}$ denotes a trajectory of observations and action
targets recorded at a fixed control frequency. AdaTempo transforms
$\mathcal{D}$ into an accelerated dataset $\mathcal{D}_{\mathrm{acc}}$ while
preserving the demonstrated task ordering. As illustrated in
Fig.~\ref{fig:pipeline}, the method first expresses tempo relatively within
each demonstration and aligns the resulting profiles to a canonical phase. It
then converts their cross-demonstration consensus into a continuous speedup
profile, which is used
to resample the demonstrations before standard policy training.

\subsection{Phase-Aligned Relative Tempo}
\label{sec:relative-tempo}

\paragraph{Motion progress and canonical reference}
For each demonstration, we first Gaussian-smooth the continuous robot-target
channels along time. From these smoothed targets, we
construct a progress descriptor $u_{i,t}$ and an alignment descriptor
$y_{i,t}$. The progress descriptor retains the continuous arm-motion
components, while the alignment descriptor additionally includes event
landmarks such as the gripper state. Each continuous motion channel is
normalized by a robust dataset-level percentile range.

We parameterize an episode by its normalized cumulative motion arc length,
\begin{equation}
\ell_i(t)=
\frac{\sum_{k=1}^{t}\lVert u_{i,k}-u_{i,k-1}\rVert_2}
{\sum_{k=1}^{T_i-1}\lVert u_{i,k}-u_{i,k-1}\rVert_2},
\qquad \ell_i(t)\in[0,1].
\label{eq:motion-progress}
\end{equation}
Unlike elapsed time, $\ell_i(t)$ measures the fraction of cumulative
demonstrated motion completed. We interpolate $y_i$ onto a common grid of
$\ell$ and choose the trajectory with the smallest median root-mean-square
distance to the others as the canonical medoid, indexed by $i^\star$.

\paragraph{Progress-regularized monotonic alignment}
Arc length provides only a coarse temporal ordering: equal values of $\ell$ need not
denote corresponding task phases. We therefore align each trajectory to the
medoid using dynamic time warping (DTW)~\cite{sakoe1978dtw}, with local cost
\begin{equation}
d_i(t,s)=\lVert y_{i,t}-y_{i^\star,s}\rVert_2^2
+\lambda_p\lvert\ell_i(t)-\ell_{i^\star}(s)\rvert^2,
\label{eq:dtw-cost}
\end{equation}
and monotonic path
\begin{equation}
W_i^*=\arg\min_{W\in\mathcal{W}}
\left[\sum_{(t,s)\in W}d_i(t,s)+\eta_i N_{\mathrm{hv}}(W)\right].
\label{eq:dtw}
\end{equation}
Here, $\mathcal{W}$ permits diagonal, horizontal, and vertical steps, and
$N_{\mathrm{hv}}(W)$ counts the non-diagonal steps penalized by $\eta_i$.
The progress term discourages implausible matches between distant parts of the task
without forcing equal timing. For each timestep $t$, we take the median
reference motion progress $\ell_{i^\star}(s)$ over all corresponding matches
$(t,s)\in W_i^*$, project the resulting sequence to be nondecreasing, and
affinely normalize its endpoints to 0 and 1. This yields the final aligned canonical
phase $p_i(t)\in[0,1]$ used below.

\paragraph{Episode-relative tempo}
We divide the canonical phase axis into $B$ equal bins. Let $n_{i,b}$ be the
occupancy of phase bin $b$ after Gaussian-smoothing its raw timestep count
across neighboring bins. A phase traversed quickly occupies
fewer timesteps, so its local tempo is proportional to the inverse occupancy,
$v_{i,b}=1/(n_{i,b}+\epsilon)$. Rather than comparing these values directly
across demonstrations, we rank them within each episode:
\begin{equation}
    r_{i,b}=
    \frac{\operatorname{rank}_{i}(v_{i,b})-1}{B-1},
    \qquad r_{i,b}\in[0,1].
    \label{eq:relative-tempo}
\end{equation}
Larger values therefore indicate phases that are relatively faster within the
same demonstration. This episode-internal normalization removes the global
tempo scale while retaining how tempo varies over canonical phase. Interpolating
the bin-wise values over canonical phase yields the aligned relative-tempo
profiles $\{r_i(p)\}_{i=1}^{N}$.

\subsection{Consensus-Guided Continuous Acceleration}
\label{sec:speed-target}

\paragraph{Tempo consensus and confidence}
To reduce sensitivity to demonstration-specific timing variations, we aggregate aligned relative-tempo profiles across demonstrations. At each phase bin, we define the cross-demonstration median as the tempo consensus and quantify disagreement using the median absolute deviation (MAD):
\begin{equation}
    m_b=\underset{i}{\operatorname{median}}\;r_{i,b},
    \qquad
    d_b=\underset{i}{\operatorname{median}}\;
    \lvert r_{i,b}-m_b\rvert.
    \label{eq:consensus-mad}
\end{equation}
We convert the dispersion into a bounded confidence
\begin{equation}
    q_b=\frac{1}{1+(d_b/\sigma_d)^2},
    \label{eq:confidence}
\end{equation}
where $\sigma_d$ controls the sensitivity to cross-demonstration disagreement.
The confidence $q_b$ measures the repeatability of the relative-tempo tendency,
rather than task success; low confidence therefore suppresses potentially unreliable
phase-wise acceleration.

\paragraph{Continuous speedup mapping}
Unlike discrete downsampling ratios, continuous speed factors allow finer-grained adjustment of acceleration across task phases. We interpolate the bin-wise consensus and confidence values
$\{m_b,q_b\}_{b=1}^{B}$ over canonical phase to obtain continuous profiles
$m(p)$ and $q(p)$. Evaluating them at each frame's aligned phase gives the
initial speedup profile
\begin{equation}
    c_i(t)
    =
    c_{\min}
    +
    \left(c_{\max}-c_{\min}\right)
    m\!\left(p_i(t)\right)
    q\!\left(p_i(t)\right).
    \label{eq:speed-mapping}
\end{equation}
Here, $c_{\min}$ and $c_{\max}$ set the task-specific speedup range, while $m(p)\,q(p)$ determines its phase-wise variation. Near the trajectory boundaries, where alignment and occupancy estimates are
less reliable, we set $c_i(t)=c_{\min}$.

A high speedup immediately before a cautious phase may over-compress the
transition into that phase. To anticipate such phases, we apply a lookahead
soft minimum over a forward window of $L$ source frames:
\begin{equation}
    \hat c_i(t)=
    -\beta\log\left[
    \frac{1}{L}\sum_{k=0}^{L-1}
    \exp\left(
    -\frac{c_i(t+k)}{\beta}
    \right)
    \right],
    \label{eq:lookahead}
\end{equation}
where indices beyond the final frame are clamped to \(T_i-1\).

We then limit frame-to-frame changes in the speedup profile:
\begin{equation}
    \begin{aligned}
    c_i^{\mathrm{target}}(t)
    ={}&c_i^{\mathrm{target}}(t-1)\\
    &+\operatorname{clip}\!\left(
    \hat c_i(t)-c_i^{\mathrm{target}}(t-1),
    -\delta_{\downarrow},\delta_{\uparrow}
    \right).
    \end{aligned}
    \label{eq:slew}
\end{equation}
We use $\delta_{\downarrow}>\delta_{\uparrow}$, allowing the profile to reduce
its speedup more rapidly than it increases it. Together, lookahead and
asymmetric slew limiting produce a smooth continuous speedup profile
that anticipates cautious phases without abrupt changes. This profile is then used for the offline trajectory resampling described below.
All remaining hyperparameters are fixed across tasks and reported in
Sec.~\ref{sec:setup}.

\subsection{Offline Resampling and Policy Training}
\label{sec:resampling}

Given $c_i^{\mathrm{target}}$, we construct one accelerated trajectory from
each source demonstration. Let $s_{i,k}$ denote the continuous position on the
source timeline corresponding to accelerated step $k$. Starting from
$s_{i,0}=0$, we advance virtual source time until reaching the final frame:
\begin{equation}
    s_{i,k+1}=\min\!\left\{
    s_{i,k}+c_i^{\mathrm{target}}(s_{i,k}),\;T_i-1
    \right\},
\label{eq:virtual-resampling}
\end{equation}
where $c_i^{\mathrm{target}}(s)$ is linearly interpolated between source
timesteps. Unlike constant-rate downsampling, Eq.~\eqref{eq:virtual-resampling}
allocates denser samples to relatively cautious phases and larger temporal
steps to phases with greater acceleration potential.

Since $s_{i,k}$ is generally non-integer, continuous quantities are evaluated
by interpolation. Specifically, for a continuous trajectory quantity $x_t$,
\begin{equation}
x(s)=(1-\alpha)x_{\lfloor s\rfloor}
+\alpha x_{\lceil s\rceil},
\qquad
\alpha=s-\lfloor s\rfloor .
\label{eq:trajectory-interpolation}
\end{equation}
At accelerated step $k$, the observation is sampled at $s_{i,k}$ and the action
target at $s_{i,k+1}$. The policy therefore learns to map the current state to
the configuration reached one accelerated control step later in the source
demonstration. Continuous joint or Cartesian quantities follow
Eq.~\eqref{eq:trajectory-interpolation}; rotations use spherical linear
interpolation (SLERP) on $\mathrm{SO}(3)$, and images use the nearest source
frame. These resampled trajectories preserve the demonstrated task order and original control frequency, and collectively form the accelerated dataset $\mathcal{D}_{\mathrm{acc}}$, which is used to train standard visuomotor policies such as ACT or DP. Because the resampled targets span
larger motions per control step, accurate low-level tracking becomes important,
as also observed in prior acceleration systems~\cite{guo2025demospeedup,arachchige2025sail}. We therefore use higher-gain low-level control to improve target tracking. All phase alignment, tempo estimation, and resampling are
performed offline. Deployment follows the trained policy's standard inference
procedure, with no additional tempo-selection module or online retiming.

\section{Experiments}
\label{sec:experiments}

We evaluate AdaTempo along four questions:
(1) can it accelerate visuomotor policies across different policy backbones
and simulated and real-world settings;
(2) how does its success--speed trade-off compare with uniform and adaptive
acceleration baselines;
(3) do demonstrations exhibit consistent relative-tempo structure within
tasks, and is this structure reproducible across demonstrators; and
(4) which components are important for reliable acceleration?

\newcommand{\resultpending}{\textcolor{gray}{--}}

\subsection{Experimental Setup}
\label{sec:setup}

\begin{figure*}[t]
    \centering
    \includegraphics[width=\textwidth]{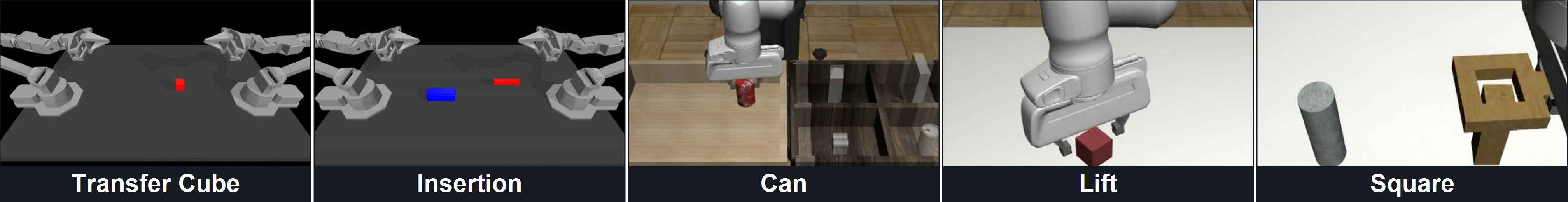}
    \caption{\textbf{Simulation tasks.} We evaluate five tasks from ALOHA and Robomimic. Transfer Cube and Insertion emphasize coordinated bimanual manipulation, while Can, Lift, and Square cover single-arm grasping and precise object manipulation.}
    \label{fig:simulation-tasks}
\end{figure*}

\paragraph{Tasks and policy backbones}
Our simulation suite contains five tasks, shown in
Fig.~\ref{fig:simulation-tasks}. For the ALOHA MuJoCo benchmark, we evaluate Transfer Cube and Insertion with ACT using the 50 human demonstrations available for each task. All ALOHA ACT policies predict 14-D absolute
joint-position targets (7 per arm, including the gripper). Can, Lift, and Square are drawn
from robomimic~\cite{mandlekar2021matters} and use DP~\cite{chi2023diffusion}, with 200 successful human demonstrations
per task. DP predicts absolute end-effector poses in Cartesian space.

\begin{figure*}[!b]
    \centering
    \includegraphics[width=\textwidth]{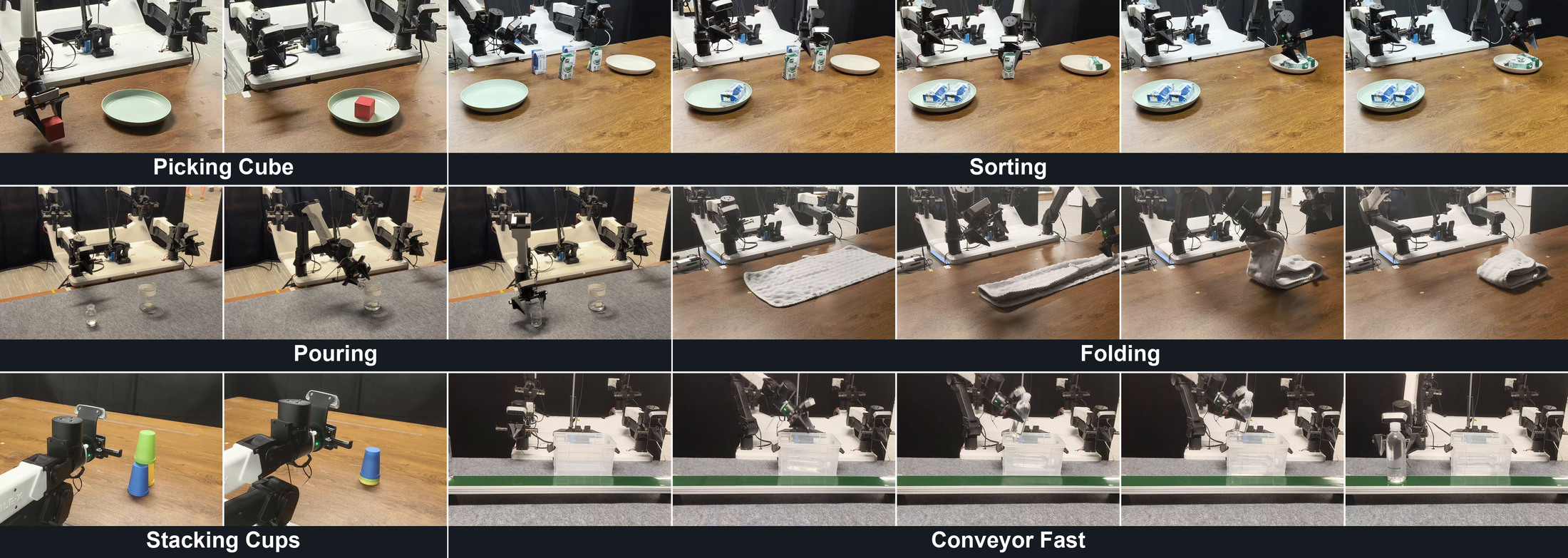}
    \caption{\textbf{Real-world evaluation tasks.} We consider six tasks covering complementary manipulation challenges. Among them, Sorting and Folding emphasize long-horizon, multi-stage manipulation; Pouring and Stacking Cups require precise coordination and spatial alignment; and Conveyor Fast stresses time-critical manipulation, requiring the robot to keep pace with fast-moving objects.}
    \label{fig:real-world-rollouts}
\end{figure*}

The real-world suite contains six manipulation tasks on an AgileX ALOHA bimanual platform
equipped with two 6-DoF PiPER follower arms: Picking Cube, Sorting, Pouring, Folding,
Stacking Cups, and Conveyor Fast, as shown in
Fig.~\ref{fig:real-world-rollouts}. For each task, we collect approximately 50 teleoperated demonstrations
(60 for Conveyor Fast) and train a separate ACT policy using RGB observations
from two wrist-mounted cameras and one overhead camera, together with
robot proprioception.

\paragraph{Compared methods}
We compare AdaTempo with the original policy, a uniform acceleration baseline
(ACT-$2\times$ or DP-Fast~\cite{arachchige2025sail}), and two representative
adaptive acceleration methods: DemoSpeedup~\cite{guo2025demospeedup} and
SAIL~\cite{arachchige2025sail}. At evaluation time, ACT‑$2\times$ temporally downsamples the original ACT action chunk by $2\times$ without retraining the policy, whereas DP-Fast executes predicted trajectories at
a higher fixed command rate. DemoSpeedup provides the closest offline
resampling baseline, while SAIL represents adaptive runtime acceleration and
uses its prescribed time-varying execution intervals and action scheduling.

In simulation, we compare all baselines with the corresponding ACT or DP
backbone. Our real-world experiments use ACT and compare AdaTempo with the
original ACT policy and DemoSpeedup; SAIL is evaluated only in simulation.
All learned methods use the same source demonstrations. Results for DP-Fast
and SAIL (DP) are taken from SAIL~\cite{arachchige2025sail}.

\paragraph{Implementation and acceleration parameters}
All AdaTempo hyperparameters are shared across tasks except the acceleration
bounds $(c_{\min},c_{\max})$. For each task, we select them from a small set of
candidate bound pairs using held-out validation rollouts and fix the selected pair
before final evaluation. Table~\ref{tab:acceleration-bounds} reports the values
used in all experiments.

We use $B=160$ canonical-phase bins and a 160-point arc-length grid for medoid
selection. Continuous robot targets and phase occupancy are Gaussian-smoothed
with standard deviations of $1.25$ frames and $3$ bins, respectively, and each
continuous motion channel is normalized by its dataset-level $5$th--$95$th
percentile range. We set $\lambda_p=0.17$,
$\eta_i=0.01\operatorname{median}_{t,s}d_i(t,s)$,
$\sigma_d=0.15$, $L=10$, $\beta=0.10$, and
$(\delta_{\uparrow},\delta_{\downarrow})=(0.06,0.15)$.
Gripper-state channels are weighted by $1.2$ during alignment.
We set $c_i(t)=c_{\min}$ over the first and last $3.5\%$ of each trajectory.
ALOHA simulation and robomimic run at nominal control rates of $50$ and
$20$~Hz, respectively. The original ACT baseline uses an action chunk size of $K=50$, while ACT
retrained on AdaTempo-accelerated trajectories uses $K=25$ to approximately
preserve the motion span of each action chunk. All ACT policies are queried
at every control step with temporal ensembling over overlapping chunks. Our DP models predict 16-step action sequences and replan every 8 steps.
For fairness, AdaTempo, DemoSpeedup, and SAIL all use
high-gain position control in ALOHA. In robomimic, AdaTempo
and DemoSpeedup both use OSC\_POSE with $K_p=3000$ and a damping
ratio of $0.5$. The real-world system operates at an effective control rate of approximately \(15\) Hz for both demonstration collection and policy deployment.

\begin{table}[t]
    \caption{Task-specific acceleration bounds.}
    \label{tab:acceleration-bounds}
    \centering
    \footnotesize
    \setlength{\tabcolsep}{8pt}
    \renewcommand{\arraystretch}{0.95}
    \begin{tabular}{lcc}
        \toprule
        Task & $c_{\min}$ & $c_{\max}$ \\
        \midrule
        Transfer Cube & 1.85 & 3.80 \\
        Insertion & 1.90 & 4.00 \\
        Lift & 3.40 & 6.50 \\
        Can & 3.40 & 6.50 \\
        Square & 1.60 & 2.40 \\
        \midrule
        Picking Cube & 1.90 & 3.80 \\
        Sorting & 1.80 & 4.20\\
        Pouring & 1.60 & 3.80\\
        Folding & 2.20 & 3.90\\
        Stacking Cups & 2.40 & 4.80\\
        Conveyor Fast & 2.00 & 4.00 \\
        \bottomrule
    \end{tabular}
\end{table}

\paragraph{Evaluation protocol and metrics}
For our simulation experiments, we train three independent policies with
different random seeds for each task and method, and evaluate each for
100 rollouts. We report each metric averaged over the three seeds. The metrics are task success rate and speedup-over-demonstration (SOD),
defined as
$\mathrm{SOD}=\bar{T}_{\mathrm{demo}}/\bar{T}_{\mathrm{succ}}$,
where $\bar{T}_{\mathrm{demo}}$ is the mean duration of the original
demonstrations and $\bar{T}_{\mathrm{succ}}$ is that of successful evaluation
rollouts. Thus, $\mathrm{SOD}>1$ indicates faster-than-demonstration execution.
Failed rollouts count toward the success rate but are excluded when computing SOD.

For real-world experiments, we conduct 30 trials per task and
report the number of successful trials and the mean completion time over
successful trials. Timing starts with the first commanded robot motion and ends
when the task-specific success condition is satisfied.

\subsection{Simulation Results}

Table~\ref{tab:simulation-results} reports the simulation results across the
ACT and DP task suites. Averaged over all five tasks, AdaTempo achieves the
highest success rate of $75.5\%$ with a mean SOD of $2.83\times$. DemoSpeedup
attains a comparable mean SOD of $2.86\times$, but a lower success rate of
$68.1\%$. This suggests that AdaTempo improves the success--speed
trade-off rather than maximizing execution speed alone.

The advantage is consistent across both policy backbones. On the ACT tasks,
AdaTempo achieves the highest success on both Transfer Cube and Insertion
while maintaining substantial acceleration. On the DP tasks, it preserves
perfect success on Lift at $3.28\times$ SOD, achieves the highest SOD on Can,
and the highest success on Square. Although competing methods obtain higher
speed on individual tasks, they often do so with a larger degradation in
success. Overall, AdaTempo provides a more balanced acceleration profile
across tasks and policy architectures.

\begin{table*}[t]
\centering
\caption{Simulation results on five manipulation tasks from the ALOHA
MuJoCo~\cite{zhao2023act} and robomimic~\cite{mandlekar2021matters}
benchmarks.}
\label{tab:simulation-results}
\vspace{2pt}
\begingroup
\setlength{\tabcolsep}{2.6pt}
\renewcommand{\arraystretch}{1.08}
\resizebox{\textwidth}{!}{%
\begin{tabular}{lcccc@{\hspace{10pt}}lcccccc}
\toprule
& \multicolumn{2}{c}{Transfer Cube}
& \multicolumn{2}{c}{Insertion}
&
& \multicolumn{2}{c}{Lift}
& \multicolumn{2}{c}{Can}
& \multicolumn{2}{c}{Square} \\
\cmidrule(lr){2-3}
\cmidrule(lr){4-5}
\cmidrule(lr){7-8}
\cmidrule(lr){9-10}
\cmidrule(lr){11-12}
Method
& \shortstack{success\\rate ($\uparrow$)}
& \shortstack{SOD\\($\uparrow$)}
& \shortstack{success\\rate ($\uparrow$)}
& \shortstack{SOD\\($\uparrow$)}
& Method
& \shortstack{success\\rate ($\uparrow$)}
& \shortstack{SOD\\($\uparrow$)}
& \shortstack{success\\rate ($\uparrow$)}
& \shortstack{SOD\\($\uparrow$)}
& \shortstack{success\\rate ($\uparrow$)}
& \shortstack{SOD\\($\uparrow$)} \\
\midrule
ACT
& 45.33\% & 1.36
& 23.33\% & 1.37
&
DP
& \textbf{100.00\%} & 1.45
& \textbf{95.33\%} & 1.46
& 82.00\% & 1.42 \\

ACT-2x
& 53.00\% & 2.28
& 12.00\% & 2.21
&
DP-Fast
& 95.00\% & 1.59
& 87.00\% & 2.48
& 55.00\% & 2.20 \\

ACT+DemoSpeedup
& 69.00\% & 2.59
& 18.00\% & \textbf{3.01}
&
DP+DemoSpeedup
& \textbf{100.00\%} & 2.67
& 80.00\% & 3.10
& 73.67\% & \textbf{2.94} \\

SAIL (ACT)
& 63.33\% & 2.53
& 15.33\% & 2.23
&
SAIL (DP)
& \textbf{100.00\%} & \textbf{3.98}
& 92.00\% & 3.20
& 86.00\% & 1.18 \\

ACT+AdaTempo
& \textbf{83.67\%} & \textbf{2.77}
& \textbf{25.00\%} & 2.65
&
DP+AdaTempo
& \textbf{100.00\%} & 3.28
& 82.67\% & \textbf{3.57}
& \textbf{86.33}\% & 1.90 \\

\bottomrule
\end{tabular}}
\endgroup
\end{table*}

\subsection{Real-World Results}

Table~\ref{tab:real-results} summarizes the real-world evaluation.
Across all six tasks, AdaTempo achieves an aggregate success rate of $77.8\%$,
substantially higher than ACT ($58.3\%$) and DemoSpeedup ($56.7\%$). At the same
time, AdaTempo reaches a mean task-wise completion-time speedup of
$2.02\times$ over ACT, comparable to the $1.95\times$ achieved by
DemoSpeedup. This demonstrates a substantially improved success--speed
trade-off rather than simply more aggressive execution.

AdaTempo achieves the best or tied-best success on four of the six tasks and
the shortest completion time on four. The advantage is particularly pronounced on Conveyor Fast, where timely
execution is critical,
while AdaTempo maintains high reliability at the fastest completion time.
Although DemoSpeedup performs better on some individual metrics, the overall
results show that AdaTempo more consistently preserves task success under
accelerated execution.

\begin{table*}[t]
\centering
\caption{Real-world evaluation on six manipulation tasks.}
\label{tab:real-results}
\begingroup
\setlength{\tabcolsep}{3.2pt}
\renewcommand{\arraystretch}{1.08}
\resizebox{\textwidth}{!}{%
\begin{tabular}{lcccccccccccc}
\toprule
& \multicolumn{2}{c}{Picking Cube} & \multicolumn{2}{c}{Sorting} &
\multicolumn{2}{c}{Pouring} & \multicolumn{2}{c}{Folding} &
\multicolumn{2}{c}{Stacking Cups} & \multicolumn{2}{c}{Conveyor Fast} \\
\cmidrule(lr){2-3}\cmidrule(lr){4-5}\cmidrule(lr){6-7}
\cmidrule(lr){8-9}\cmidrule(lr){10-11}\cmidrule(lr){12-13}
Method & \shortstack{success\\($\uparrow$)} & \shortstack{time (s)\\($\downarrow$)} &
\shortstack{success\\($\uparrow$)} & \shortstack{time (s)\\($\downarrow$)} &
\shortstack{success\\($\uparrow$)} & \shortstack{time (s)\\($\downarrow$)} &
\shortstack{success\\($\uparrow$)} & \shortstack{time (s)\\($\downarrow$)} &
\shortstack{success\\($\uparrow$)} & \shortstack{time (s)\\($\downarrow$)} &
\shortstack{success\\($\uparrow$)} & \shortstack{time (s)\\($\downarrow$)} \\
\midrule
ACT & 22/30 & 11.2 & 15/30 & 35.6 &
\textbf{26/30} & 32.2 & 18/30 & 30.3 &
\textbf{22/30} & 17.9 & 2/30 & 6.9 \\
ACT+DemoSpeedup & 24/30 & \textbf{5.9} & \textbf{18/30} & 19.4 &
23/30 & \textbf{17.8} & 8/30 & 15.4 &
17/30 & 8.1 & 12/30 & 3.5 \\
ACT+AdaTempo & \textbf{28/30} & 6.1 & 14/30 & \textbf{18.2} &
\textbf{26/30} & 18.3 & \textbf{23/30} & \textbf{14.2} &
20/30 & \textbf{7.8} & \textbf{29/30} & \textbf{3.2} \\
\bottomrule
\end{tabular}}
\endgroup
\end{table*}

\subsection{Relative-Tempo Consistency Across Demonstrations}
\label{sec:tempo-consistency}

\begin{figure*}[t]
    \centering
    \begin{minipage}[t]{0.355\textwidth}
        \centering
        \includegraphics[width=\linewidth]{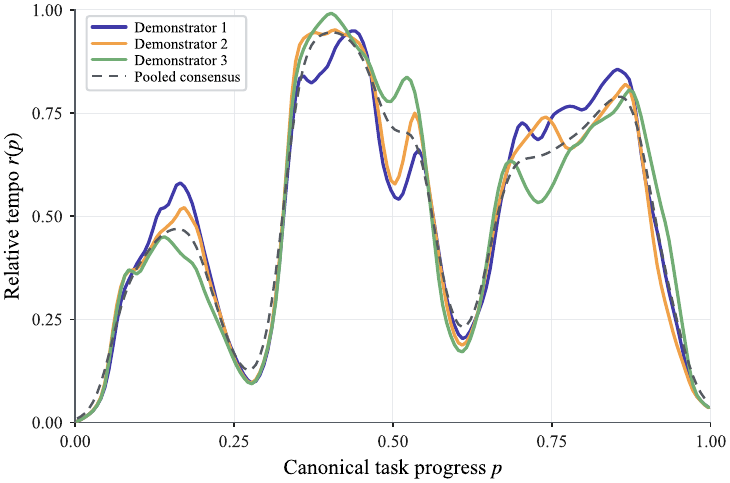}
        \vspace{-1.0mm}

        {\footnotesize\textbf{(a)} Cross-demonstrator consistency}
    \end{minipage}\hfill
    \begin{minipage}[t]{0.625\textwidth}
        \centering
        \includegraphics[width=\linewidth]{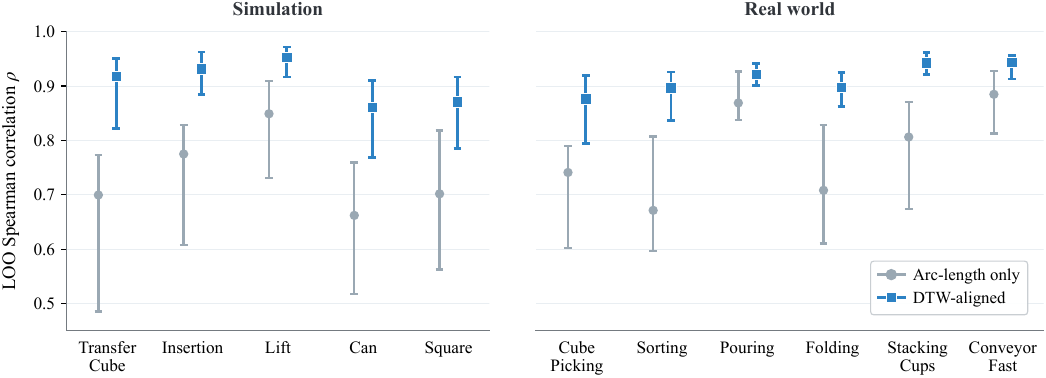}
        \vspace{-1.0mm}

        {\footnotesize\textbf{(b)} Cross-demonstration consistency}
    \end{minipage}
    \vspace{-1.0mm}
   \caption{\textbf{Relative-tempo consistency.}
(a) Demonstrator-wise consensus profiles on Conveyor Fast from three
demonstrators, together with the pooled consensus.
(b) Median and interquartile range of leave-one-out Spearman correlations
across demonstrations on all 11 tasks.}
    \label{fig:tempo-consistency}
\end{figure*}

We examine relative-tempo consistency at two levels: across demonstrators and
across demonstrations within the same task. For cross-demonstrator consistency,
three demonstrators independently collected 20 Conveyor Fast demonstrations
each. After aligning all 60 demonstrations to a common canonical phase, we
compute a consensus profile for each demonstrator. The three profiles closely follow the pooled consensus
(Fig.~\ref{fig:tempo-consistency}(a)) and have pairwise Spearman correlations
of $0.983$, $0.941$, and $0.955$, indicating that the observed relative-tempo
structure is not merely specific to one demonstrator's execution style.

We quantify cross-demonstration consistency on all 11 tasks. For episode $i$,
let $r_i\in[0,1]^B$ denote its relative-tempo profile and define the
leave-one-out consensus
\begin{equation}
    \bar r_{-i,b}
    =
    \underset{j\neq i}{\operatorname{median}}\;r_{j,b}.
\end{equation}
We compute $\rho_i=\operatorname{Spearman}(r_i,\bar r_{-i})$ to measure
agreement in the ordering of faster and slower phases. As a correspondence
baseline, \emph{arc-length only} replaces the DTW-derived phase $p_i(t)$ with
normalized cumulative motion arc length $\ell_i(t)$, while leaving subsequent
tempo computation unchanged. The DTW cost contains no tempo information.

Figure~\ref{fig:tempo-consistency}(b) reports the median and interquartile range of $\rho_i$. With arc-length-only correspondence, task-wise median
correlations range from $0.662$ to $0.885$, indicating that recurring
relative-tempo structure already exists across demonstrations. DTW alignment
raises the range to $0.861$--$0.952$ and improves every task, making the shared
structure more consistently observable through task-phase correspondence.

\subsection{Ablation Studies}
\label{sec:ablations}

We conduct focused ablation studies on Transfer Cube and Insertion, which share the
ALOHA simulator and ACT backbone but differ in precision demands. We compare
the full method with four variants.

\begin{center}
\begin{minipage}{\columnwidth}
\centering
\captionof{table}{Ablation results on the ALOHA simulation tasks.}
\label{tab:ablations}
\setlength{\tabcolsep}{3.0pt}
\renewcommand{\arraystretch}{1.08}
\resizebox{\columnwidth}{!}{%
\begin{tabular}{lcccc}
\toprule
& \multicolumn{2}{c}{Transfer Cube}
& \multicolumn{2}{c}{Insertion} \\
\cmidrule(lr){2-3}\cmidrule(lr){4-5}
Variant
& \shortstack{success\\rate ($\uparrow$)}
& \shortstack{SOD\\($\uparrow$)}
& \shortstack{success\\rate ($\uparrow$)}
& \shortstack{SOD\\($\uparrow$)} \\
\midrule
\textbf{Full AdaTempo}
& 83.67\% & 2.77
& 25.00\% & 2.65 \\
\midrule
Single-demo relative tempo
& 82.00\% & 2.57
& 20.33\% & 2.64 \\

w/o confidence weighting
& 77.00\% & 2.79
& 22.00\% & 2.99 \\

w/o temporal regularization
& 72.33\% & 2.56
& 23.67\% & 2.96 \\

RBD-style augmentation
& 83.33\% & 2.74
& 19.00\% & 2.57 \\
\bottomrule
\end{tabular}}
\end{minipage}
\end{center}

\emph{Single-demo relative tempo} replaces the cross-demonstration consensus
with each episode's own relative-tempo profile, which is mapped to
$c_i^{\mathrm{target}}$ for offline trajectory resampling.
\emph{Without confidence} sets $q(p)=1$, removing the down-weighting of phases
with high cross-demonstration tempo dispersion.
\emph{Without temporal regularization} removes the lookahead soft minimum and
asymmetric slew limits, directly using the resulting instantaneous speed curve.
\emph{RBD-style augmentation} adds offset resampling variants to improve
source-state coverage, following the replicate-before-downsample principle used
for integer-rate downsampling~\cite{guo2025demospeedup,kim2025espada}.

Table~\ref{tab:ablations} summarizes the results. Cross-demonstration consensus
improves robustness by suppressing demonstration-specific timing variations:
using a single demonstration's tempo profile reduces success, particularly on
Insertion. Confidence weighting further improves reliability by moderating
acceleration in phases where demonstrations disagree; removing it yields higher
SOD but lower success on both tasks. Temporal regularization provides an
additional stabilizing effect, with its removal producing a more aggressive
success--speed trade-off, especially on Insertion.

RBD-style augmentation provides no consistent benefit. Unlike integer-rate
downsampling, where offset replicas recover source frames omitted by a fixed
stride, continuous nonuniform resampling already samples at fractional source
times. Additional offset variants therefore introduce more interpolated
state--action targets rather than simply recovering discarded demonstrated
states, which may explain why RBD-style augmentation does not improve
performance in our experimental setting.

\section{Conclusion and Limitations}
\label{sec:conclusion}

We presented AdaTempo, an offline method that turns relative-tempo structure
shared across demonstrations into accelerated training trajectories. Standard visuomotor policies can then be trained without auxiliary models and deployed without additional runtime mechanisms. Across five simulated and six real-world tasks, AdaTempo reaches up to $3.57\times$ SOD; it achieves the highest average simulation success at a
mean $2.83\times$ SOD and, in the real world, raises aggregate success from
$58.3\%$ to $77.8\%$ over ACT while completing tasks $2.02\times$ faster on
average. These results support cross-demonstration relative tempo as a useful
self-supervised cue for phase-dependent speedup.

Relative tempo is an empirical cue rather than a direct estimate of the
maximum feasible speed. Shared teleoperation habits may influence it, and a
single medoid and consensus may be inadequate when demonstrations contain
unrelated pauses, ambiguous correspondences, or qualitatively different
strategies. Strategy-aware clustering is a natural extension for such data.

The acceleration bounds $(c_{\min},c_{\max})$ are also task-specific and are
currently selected with held-out validation rollouts. In addition, resampling
cannot remove physical limits: aggressive targets may expose tracking error,
actuator limits, latency, or unmodeled contact dynamics, while high-gain control
provides no formal safety guarantee. Finally, the learned tempo is fixed at
deployment and cannot adapt online to changing conditions. Automatic bound selection and lightweight feasibility-aware adaptation are important directions for future work.

\section*{Acknowledgment}
The authors thank the Institute of Trustworthy Embodied Artificial Intelligence,
Fudan University, for providing access to the ALOHA platform used in the
real-world experiments.

\bibliographystyle{IEEEtran}
\bibliography{references}

\end{document}